\documentclass[conference]{IEEEtran}
\IEEEoverridecommandlockouts
\usepackage{cite}
\usepackage{amsmath,amssymb,amsfonts}
\usepackage{algorithmic}
\usepackage{subfigure}
\usepackage{graphicx}
\usepackage{stfloats} 
\usepackage{textcomp}
\usepackage{multirow}
\usepackage{xcolor}
\def\BibTeX{{\rm B\kern-.05em{\sc i\kern-.025em b}\kern-.08em
    T\kern-.1667em\lower.7ex\hbox{E}\kern-.125emX}}
\makeatletter % changes the catcode of @ to 11
\newcommand{\linebreakand}{%
  \end{@IEEEauthorhalign}
  \hfill\mbox{}\par
  \mbox{}\hfill\begin{@IEEEauthorhalign}
}

\usepackage{bbold}
\usepackage{physics}
\usepackage{mathtools}
\usepackage{verbatim}
\usepackage{xr}
\usepackage{stmaryrd}

\usepackage{balance}
\usepackage{algorithm}
\usepackage{algorithmic}

\usepackage{placeins}
\usepackage{booktabs}

\makeatother % changes the catcode of @ back to 12

\usepackage{fancyhdr}
\begin{document}

\title{Active Dendrites Enable Efficient Continual Learning\\ in Time-To-First-Spike Neural Networks}
%\iffalse
\author{\IEEEauthorblockN{Lorenzo Pes$^{1,2\star}$, Rick Luiken$^{1}$, Federico Corradi$^{1}$ and Charlotte Frenkel$^{2}$}
\textrm{$^{1}$ Electrical Engineering Department, Eindhoven University of Technology, Eindhoven, The Netherlands}\\
\textrm{$^{2}$ Microelectronics Department, Delft University of Technology, Delft, The Netherlands}\\
\textit{$^{\star}$ Correspondence to l.pes@tue.nl}
}
%\fi
\maketitle
\thispagestyle{fancy}

\begin{abstract}
While the human brain efficiently adapts to new tasks from a continuous stream of information, neural network models struggle to learn from sequential information without catastrophically forgetting previously learned tasks. This limitation presents a significant hurdle in deploying edge devices in real-world scenarios where information is presented in an inherently sequential manner. Active dendrites of pyramidal neurons play an important role in the brain's ability to learn new tasks incrementally. By exploiting key properties of time-to-first-spike (TTFS) encoding and leveraging its high sparsity, we present a novel spiking neural network (SNN) model enhanced with active dendrites. Our model can efficiently mitigate catastrophic forgetting in temporally-encoded SNNs, which we demonstrate with an end-of-training accuracy across tasks of 88.3\% on the test set using the Split MNIST dataset. Furthermore, we provide a novel digital hardware architecture that paves the way for real-world deployment in edge devices. Using a Xilinx Zynq-7020 SoC FPGA, we demonstrate a 100-\% match with our quantized software model, achieving an average inference time of 37.3 ms and an 80.0\% accuracy.

%Using a Xilinx Zynq-7020 SoC FPGA, we demonstrate a 100-\% matching with our quantized software model and an average inference time of 37.3 ms. 

%\vspace*{15\baselineskip}
\end{abstract}


\section{Introduction}

\begin{figure}[t]
      \centering
      {\includegraphics[width=0.48\textwidth]{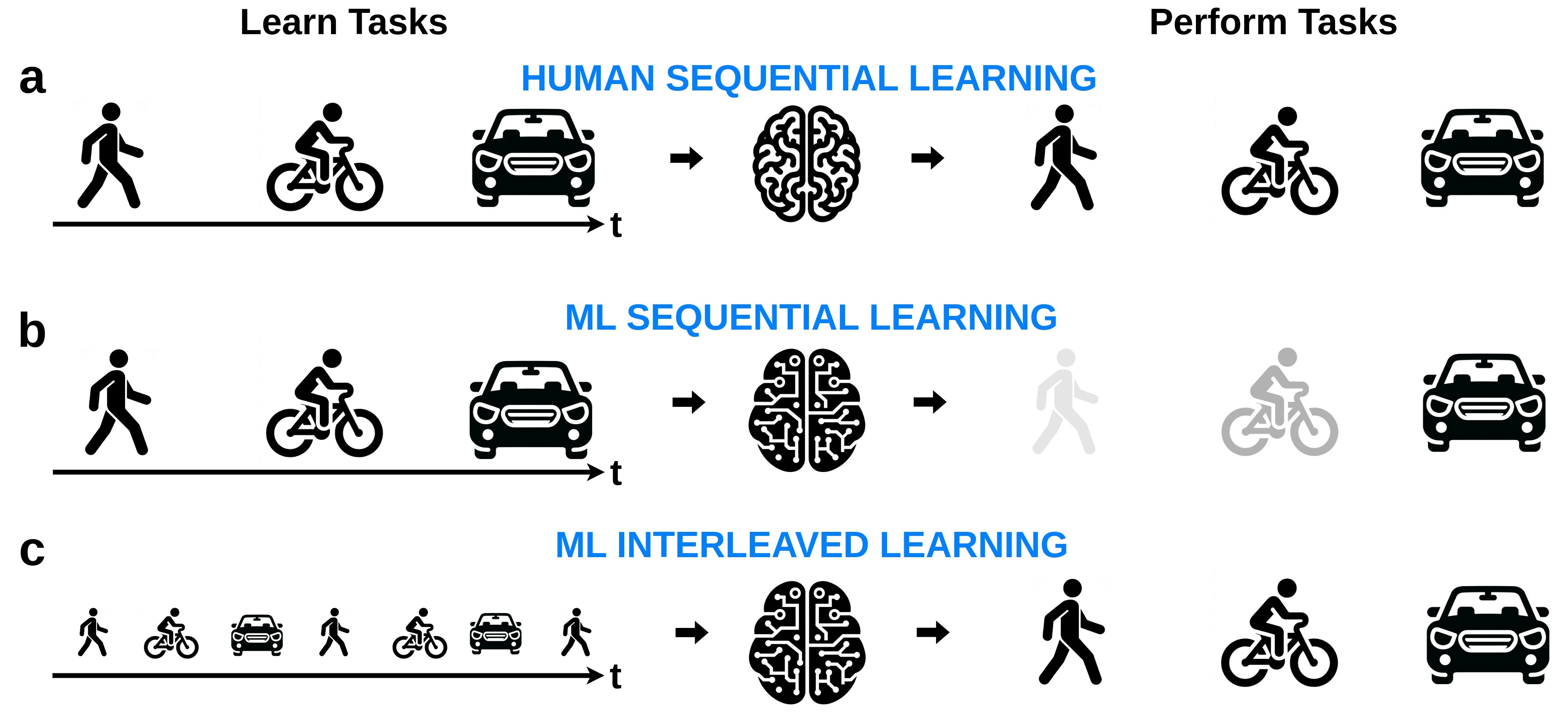}}
      \caption{\textbf{Human versus conventional ML learning}. \textbf{a} The human brain can learn new tasks in a sequential order without forgetting previous ones. \textbf{b} Training ML models in a sequential order leads to catastrophic forgetting. \textbf{c} Interleaving samples when training ML models avoids catastrophic forgetting.}
      \label{fig:fig1}
\end{figure}

As humans experience the physical world, they demonstrate the innate ability to sequentially learn new tasks without forgetting how to perform previously learned ones. For example, consider the consecutive learning experiences of a human as depicted in~\figrefsub{fig:fig1}{a}, from the initial steps of walking to the ability to drive a bike without falling, and finally to the more complex task of driving a car. Humans do not forget to walk or ride a bike when learning to drive a car. In stark contrast, as illustrated in~\figrefsub{fig:fig1}{b}, machine learning (ML) models typically struggle to learn new tasks sequentially without forgetting previously learned tasks~\cite{1989_McCloskey_catastrophic_inference}. To mitigate this problem, conventional training methodologies based on \textit{error backpropagation} (BP) \cite{1986_rumelhart_learning} and \textit{stochastic gradient descent} (SGD) \cite{1951_robbins_stochastic}, rely on examples of different tasks being presented in an interleaved fashion, as in~\figrefsub{fig:fig1}{c}. 

%it also underscores a fundamental contrast in the learning paradigm employed by humans compared to AI systems. Specifically, while humans can incrementally learn new tasks, AI requires prior knowledge of all the tasks it needs to learn, hindering its adaptability in dynamic and evolving environments
However, while this approach is at the core of today's state-of-the-art performance of ML models on pattern recognition~\cite{2019_Abiodun_PR_review}, object detection in images and videos~\cite{2012_Krizhevsky_imagenet,2019_zhao_object}, natural language processing~\cite{2020_brown_language,2013_mikolov_distributed}, and speech recognition~\cite{2019_schneider_wav2vec}, real-world scenarios at the edge mostly rely on information being presented in a sequential fashion. To deploy ML techniques in such use cases without suffering from \textit{catastrophic forgetting}~\cite{1989_McCloskey_catastrophic_inference}, various approaches have been proposed and can be categorized into three groups: \textit{regularization-based}, attempting to prevent changes in parameters that are important for a previously-learned task~\cite{2017_kirkpatrick_ewc,2017_Zenke_Continual_Learning_Synaptic_Intelligence}; \textit{architectural},
which use a subset of parameters for each new task \cite{2018_Masse_Alleviating_Catastrophic_Forgetting_EWC, 2022_iyer}; and \textit{replay-based}, where data from previous tasks is presented again to the network while new tasks are being trained \cite{2017_shin_continual,2020_van_brain}.

Beyond suffering from catastrophic forgetting, current ML systems still lag orders of magnitude behind their biological counterparts in terms of energy efficiency~\cite{2023_Frenkel}. In an attempt to better approach brain's efficiency, spiking neural networks (SNNs) are an increasingly popular network model. 
%that relies on the following two key features compared to standard ANNs: (i)~membrane potential dynamics endow SNNs with a continuous-time state, and (ii)~spike-based activation allows for sparse and event-driven communication and computation. 
Various spike coding schemes have been investigated to represent information, with the most common ones being:
\begin{itemize}
    \item \textit{Rate coding --} Information is encoded in the instantaneous frequency of spike streams. This coding scheme is popular in SNN models thanks to its robustness and ease of use, where precision can be achieved at the expense of sparsity.%This coding scheme has been observed experimentally in biological sensory and motor systems~\cite{2021_Guo_Neural_Coding_in_SNN}
    \item \textit{Time-to-first-spike (TTFS) coding --} Information is encoded into the spike time from an initial observation reference, where the more important the information, the earlier the spike. Often combined with the assumption that each neuron spikes at most once, TTFS coding outlines significant energy savings in SNN hardware as it allows optimizing for sparsity~\cite{2020_frenkel}.%This coding scheme has been hypothesized to explain the ability of the bain to provide a fast and efficient response~\cite{2011_Ponulak_intro_to_snn}.
\end{itemize}

Nevertheless, TTFS-based networks are typically considered difficult to train due to the \textit{dead neurons} problem~\cite{2021_zhang}, where inactive neurons do not contribute to the learning process. This issue, combined with catastrophic forgetting, is currently hindering the adoption of TTFS-encoded SNNs for adaptive edge computing based on streaming data. In this work, we solve this challenge by introducing the concept of \textit{active dendrites}~\cite{2022_iyer} in TTFS-encoded SNNs. Active dendrites, coupled with a gating mechanism, allow for a dynamic selection of different sub-networks for different tasks, which mitigates catastrophic forgetting by avoiding overwriting previous knowledge. Interestingly, the dead neurons problem of TTFS networks can be exploited to perform this gating mechanism intrinsically. We demonstrate these findings by showcasing a test accuracy of 88.3\% in sequentially training tasks based on the Split MNIST dataset. Additionally, we propose a digital hardware architecture for TTFS-encoded SNNs enhanced with active dendrites, which can perform inference with an average time of 37.3 ms while fully matching the results from our quantized software model.

\section{Background material}
\subsection{Time-to-first-spike neural networks}
\label{sub:srm_with_rel_psp_kernel}

In light of their compelling performance in terms of inference time and power consumption, TTFS neural networks sparked a strong interest in the field of \textit{neuromorphic computing} \cite{2002_Bohtea_BP_in_Temporal_Encoded_NN,2017_Mostafa_Supervised_Learning_For_Temporal_Coding, 2020_kheradpisheh_temporal,2021_comcsa_temporal}. While most previous approaches encounter problems of non-differentiable spike functions \cite{2002_Bohtea_BP_in_Temporal_Encoded_NN} and exploding gradients \cite{2017_Mostafa_Supervised_Learning_For_Temporal_Coding,2020_kheradpisheh_temporal}, Zhang \textit{et al.} in \cite{2021_zhang} propose an elegant solution to these issues by defining the membrane potential evolution of a neuron as

\begin{equation}
    V_j^l(t)= 
    \begin{cases}
        \sum_i^I W^l_{ij} (t - t^{l-1}_i), \;\;\; \text{if} \; t>t^{l-1}_i\\
        0, \;\;\;\;\;\;\;\;\;\;\;\;\;\;\;\;\;\;\;\;\;\;\;\;\;\;\;\;\text{otherwise},
    \end{cases}
    \label{eq:relpsp_potential}
\end{equation}
\noindent
where $V_{j}^l$ is the membrane potential of neuron $j$ in layer $l$, $t$ is the simulation timestep, $t^{l-1}_i$ is the spike time of the pre-synaptic neuron $i$ in the preceding layer $l-1$ and $W^l_{ij}$ is the synaptic strength connecting neuron $i$ to neuron $j$. Specifically, $i \in \mathbb{R^I}$ and $j \in \mathbb{R^J}$, with $I$ being the total number of neurons in layer $l-1$ and $J$ the total number of neurons in layer $l$.

When a pre-synaptic neuron $i$ emits a spike, the membrane potential of the post-neuron $j$ linearly integrates the synaptic connection $W^l_{ij}$ as depicted in~\figrefsub{fig:fig2}{a}. During the integration process, if the membrane potential of neuron $j$ crosses the \textit{threshold voltage}, i.e $V_{th}$, a spike is emitted as

\begin{equation}
    S_j^l(t)= 
    \begin{cases}
        1, \;\;\; \text{if} \; V_j^l(t) \geq V_{th}\\
        0, \;\;\;\text{otherwise}.
    \end{cases}
    \label{eq:spike_eq}
\end{equation}
\noindent
The time at which this event occurs, that is, $t^l_{j}$, is called the \textit{spike time} and can be found by plugging $V^l_{j}(t_j) = V_{th}$ into (\ref{eq:relpsp_potential}), resulting in

\begin{equation}
    t^l_j = \frac{V_{th} + \sum_i^I W^l_{ij} t^l_i}{\sum_i^I W^l_{ij}}, \;\;\; \text{if} \; t>t^{l-1}_i.
    \label{eq:rel_psp_spike_time}
\end{equation}

If the spike time of a neuron occurs after the maximum observation windows defined by $T_{max}$, the neuron is considered \textit{dead}, carrying no information and consequently having a gradient of zero during error BP. According to Zhang \textit{et al.} in~\cite{2021_zhang}, the required gradients for applying error backpropagation are

\begin{equation}
 \frac{\partial t^l_j}{\partial W^l_{ij} } =   \frac{\partial t^l_j}{\partial V(t^l_j)} \frac{\partial V(t^l_j)}{\partial W^l_{ij}} = \frac{t_i^{l-1} - t^l_{j}}{\sum_i^I W^l_{ij}},
 \label{eq:w_update_rel_psp}
\end{equation}

\begin{equation}
 \frac{\partial t^l_j}{\partial t^{l-1}_i } =   \frac{\partial t^l_j}{\partial V(t^l_j)} \frac{\partial V(t^l_j)}{\partial t^{l-1}_i} = \frac{W^l_{ij}}{\sum_i^I W^l_{ij}}. 
 \label{eq:downstream_update_rel_psp}
\end{equation}
\noindent
%Notably, as the membrane potential grows linearly at the spike time, the derivative $\partial t^l_j/\partial V(t^l_j)$ can be directly calculated and not approximated as in \cite{2002_Bohtea_BP_in_Temporal_Encoded_NN}. 

\label{bm:rel_psp_kernel}
\subsection{Active dendrites in artificial neural networks}
\label{bm:active_dendrites}

\begin{figure}[t]
  \centering
  {\includegraphics[width=0.48\textwidth]{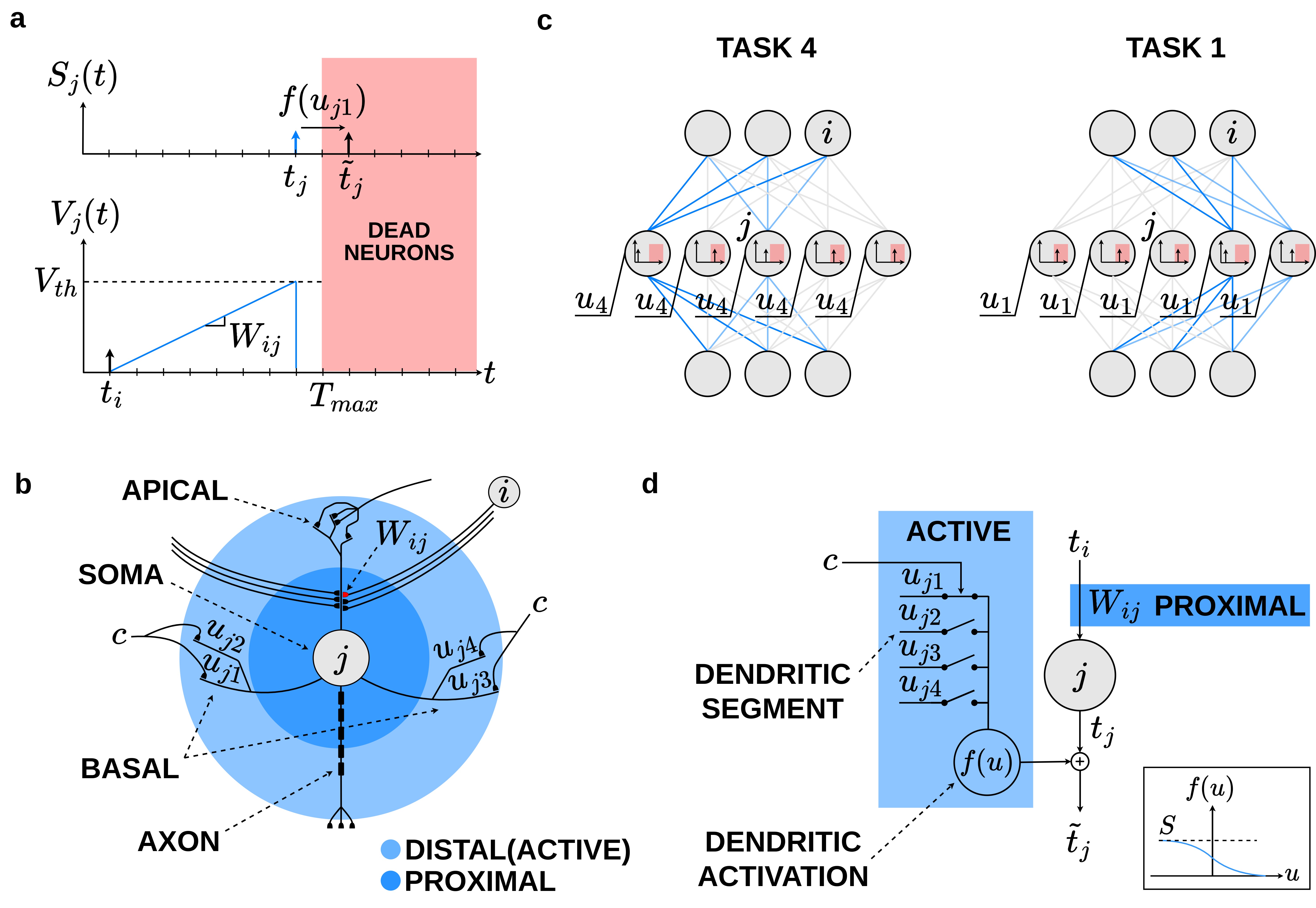}}
  \caption{\textbf{Neuron model and network architecture.} \textbf{a} Bottom: linear integration of synaptic strength $W_{ij}$ following a pre-synaptic spike at $t_i$. Top: dendritic modulation of the spike time delay.  \textbf{b} Illustration of a pyramidal neuron. \textbf{c} Selection of different sub-networks for different tasks based on the activity of dendritic segments. Dead neurons are used to efficiently implement a gating mechanism, with outgoing synaptic connections shown in gray (connections from active neurons are shown in blue). \textbf{d} Proposed neuron model and dendritic activation function. }
  \label{fig:fig2}
\end{figure}

Conventional artificial and spiking neuron models are historically rooted in the \textit{point neuron} model, which assumes a linear impact of all synapses on the membrane potential~\cite{1907_Lapicque_point_neuron}. This model lacks the architectural organization and dynamics found in the most abundant neurons in the cerebral cortex: pyramidal neurons, illustrated in \figrefsub{fig:fig2}{b}. Synaptic connections of a neuron are located on dendritic branches, which are referred to as proximal or distal when they are close or far from the neuron body (soma), respectively. Furthermore, dendrites can be of two types: basal or apical, depending on whether they connect to the soma or the apex of the neuron.
While synapses located in proximal dendrites are believed to linearly scale the input signal of neighboring neurons, distal dendrites, also referred to as \textit{active dendrites}, perform non-linear local integration of the input signals \cite{2016_hawkins_neurons}. Specifically, basal active dendrites process contextual information from neighboring cortical areas \cite{2000_yoshimura_properties} and modulate the activity of the soma in a context-dependent manner \cite{2020_takahashi_active}. 

%while apical active dendrites process feedback signals \cite{2008_spruston_pyramidal}. 

Inspired by these ideas, Iyer \textit{et al.}~in \cite{2022_iyer} demonstrate that enhancing an artificial neuron model with the processing capabilities of active dendrites mitigates the problem of catastrophic forgetting by fostering the emergence of different subnetworks for different tasks. Specifically, the authors introduce a modulation effect on the feedforward activation $y$ of a neuron $j$ to mimic the behavior of active dendrites as 

\begin{equation}
    \tilde{y_j} = y_j \times 
     \sigma(\underset{j}{\text{max}} u^T_jc),
\label{eq:modulated_neuron_iyer}
\end{equation}

\noindent
where, $\tilde{y_j}$ is the modulated activation, $\sigma(\cdot)$ is a sigmoidal dendritic activation function, $u_j$ is a dendritic segment and $c$ is the \textit{context vector}. In a continual learning setup, the context vector is changed upon the start of a new task and a different dendritic segment is selected accordingly. Based on the value of the selected segment, the activity of a neuron can be down-modulated or remain unchanged.
To foster the emergence of different subnetworks for different tasks, Iyer et al.~\cite{2022_iyer} propose to add a \textit{k}WTA layer after each layer enhanced with active dendrites. This layer selects the \textit{k} neurons with the highest activation and gates the others. In this manner, only a small subset of neurons in the layer is activated for a given task. Furthermore, during the backward pass, only the synaptic strengths and the active dendritic segment of the winning neurons are updated. By following this approach, different subnetworks emerge for each task, thereby reducing interference between tasks and mitigating catastrophic forgetting.

\section{Proposed active-dendrites algorithmic and hardware framework for TTFS-encoded networks}

\subsection{Neuron model and network architecture}
\label{sub:proposed_nn}
To mitigate the problem of catastrophic forgetting in SNNs while exploiting the sparsity of TTFS encoding, our neuron model enhances the model of \cite{2021_zhang} with a simplified version of the active dendrites proposed in~\cite{2022_iyer}. As explained in Section~\ref{bm:active_dendrites}, active dendrites modulate the activity of artificial neurons in a context-dependent manner. However, while artificial neurons encode information in real-valued numbers, TTFS neurons encode information in the spike time. Thus, to modulate the activity of a TTFS neuron, we introduce a dendritic-dependent spike time delay mechanism as follows: 

\begin{equation}
    \tilde{t^l_j} = t^l_j + f(u^l_{j n}), 
    \label{eq:catastrosolve_spike_Time}    
\end{equation}
\noindent
where $\tilde{t^l_j}$ is the dendritic-modulated spike time, $t^l_j$ is the spike time defined in Eq.~(\ref{eq:rel_psp_spike_time}), $u^l_{jn}$ is the dendritic segment selected for task $n$, and $f(\cdot)$ is the dendritic activation function defined as

\begin{equation}
    f(u)= {\frac{S}{1+e^{u}}},
    \label{eq:activation_func}
\end{equation}
\noindent
where $S$ is a hyperparameter that controls the strength of the dendritic delay. It ensures that negative dendrites increase delay, while positive dendrites reduce delay.

The minimum number of dendritic segments for each neuron must match the number of tasks, that is, $u_{j} \in \mathbb{R^N}$, where $N$ is the total number of tasks. Depending on the task being performed, the context vector $c$ connects a different dendritic segment to the activation function. As illustrated in \figrefsub{fig:fig2}{c}, if task $n = 4$ is performed, dendrite $u_{4}$ is connected to the activation function. Using this approach, we generate a similar context-dependent modulation effect as the one proposed in \cite{2022_iyer} and expressed in Eq.~(\ref{eq:modulated_neuron_iyer}).

However, as opposed to the approach from Iyer et al. that necessitates a dedicated \textit{k}WTA layer, we can exploit dead neurons in our TTFS-encoded network to intrinsically implement a gating mechanism. Indeed, a dead neuron is equivalent to a real value of zero in a network of artificial neurons, acting as a gating mechanism similar to a \textit{k}WTA layer. Moreover, the dendritic-dependent spike time delay mechanism can push neurons that otherwise would have been active in the dead zone, thereby forming a dynamic context-dependent gating mechanism.  A pictorial representation of the proposed neuron model and the activation function is provided in \figrefsub{fig:fig2}{d}.

Following the introduction of active dendrites, the model contains two learnable parameters $\Theta = (W^l,u^l)$, which are modified by BP to minimize a spike-time-based cross-entropy loss function as proposed in \cite{2021_zhang}. With the new model equations, the gradients expressed in Section~\ref{sub:srm_with_rel_psp_kernel} thus become

\begin{equation}
 \frac{\partial t^l_j}{\partial W^l_{ij} } = \frac{\partial t^l_j}{\partial V(t^l_j)} \frac{\partial V(t^l_j)}{\partial W^l_{ij}} = \frac{t_i^{l-1} - t^l_{j} + f(u^l_{jn})}{\sum_i^I W^l_{ij}},
 \label{eq:w_update_catastrosolve}
\end{equation}

\begin{equation}
 \frac{\partial t^l_j}{\partial u^l_{jn} } = \frac{\partial t^l_j}{\partial V(t^l_j)} \frac{\partial V(t^l_j)}{\partial u^l_{jn}} = \frac{f'(u^l_{jn})}{\sum_i^I W^l_{ij}},
 \label{eq:u_update_catastrosolve}
\end{equation}

\noindent
where Eq.~(\ref{eq:u_update_catastrosolve}) defines the direction of steepest descent for the dendritic segment $u^l_{jn}$. Note that the downstream gradient expressed in Eq.~(\ref{eq:downstream_update_rel_psp}) remains unaffected by the introduction of active dendrites. 

\subsection{Digital hardware architecture}
\begin{figure}[t!]
  \centering
  {\includegraphics[width=0.48\textwidth]{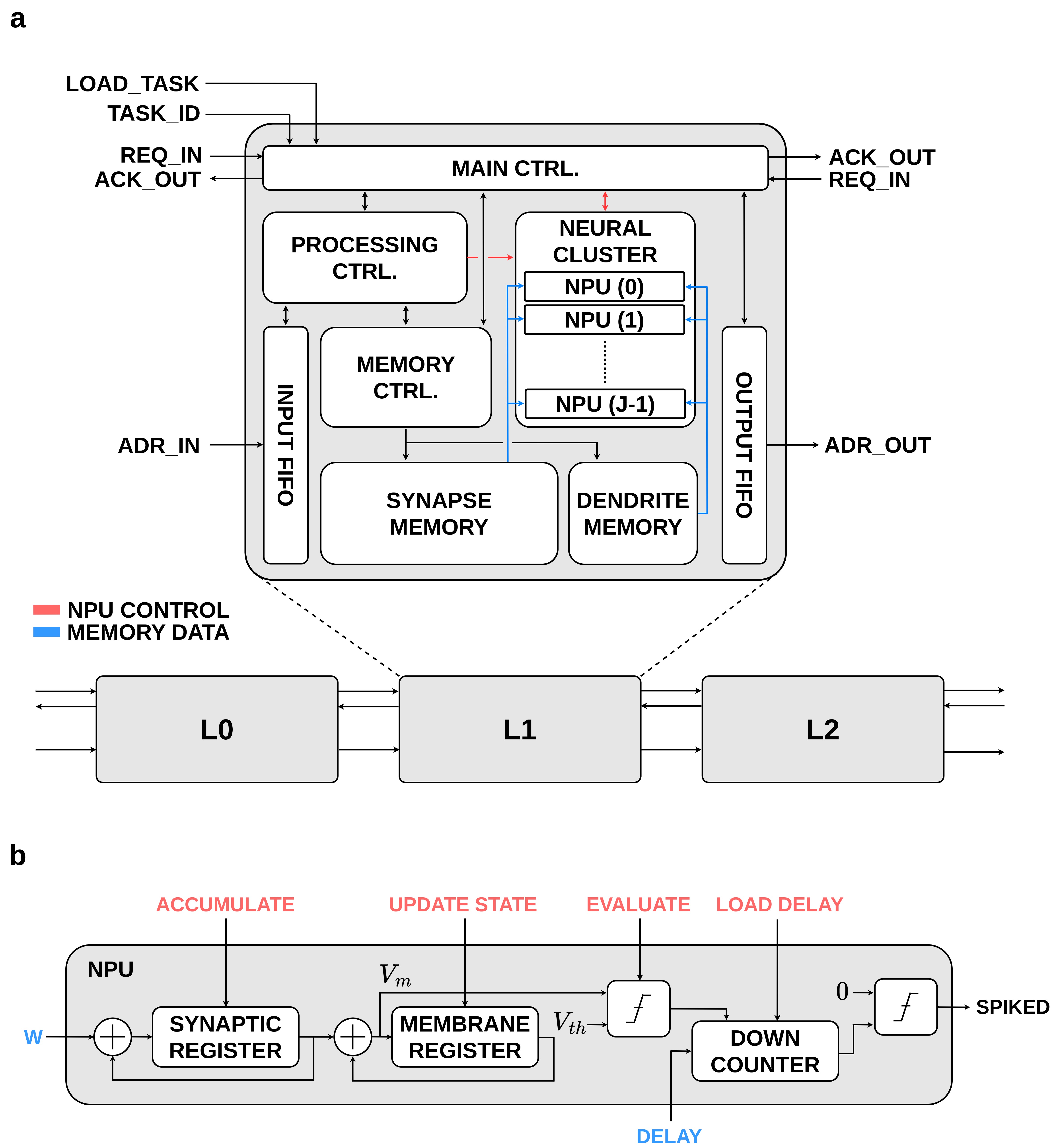}}
  \caption{\textbf{Digital Hardware Architecture. a} Architecture of a layer containing $J$ parallel neuron processing units (NPUs). \textbf{b} Block diagram of an NPU implementing the dynamics of the proposed neuron model. The synaptic register accumulates the synaptic strength (i.e., $W$) received at each timestep. The membrane register stores the membrane voltage of the neuron. It integrates the value stored in the synaptic register at the end of each timestep. The down counter is loaded with the dendritic delay of the current task. Following a threshold crossing event, it starts decrementing at each timestep. When its value reaches zero, it raises the \textit{SPIKED} flag. }

  \label{fig:fig3}
\end{figure}

The proposed architecture is inspired by Gyro \cite{2021_corradi_gyro}, a digital event-driven architecture supporting multiple fully-connected layers of spiking neurons, as depicted in \figrefsub{fig:fig3}{a} for three layers. Each layer of Gyro consists of a \textit{memory module} storing synaptic connections between adjacent layers, a \textit{neural cluster} containing parallel instances of the neuron processing unit (NPU), \textit{input/output queues} storing the address of spiking pre-synaptic and post-synaptic neurons, and \textit{control modules} implemented as finite-state-machines (FSMs) controlling the communication between layers and the interaction between modules within a layer.

Adjacent layers communicate through a 4-phase handshake protocol orchestrated by the main controller. At each timestep, the address of spiking pre-synaptic neurons is pushed to the input queue of the succeeding layer, followed by a request signal to initiate processing. In response to this request, the processing controller pops the addresses and triggers the memory controller to access the synapse memory at the received locations. The addresses of the synaptic memory are aligned with the indices of pre-synaptic neurons and each address contains all the synaptic weights connecting a pre-synaptic neuron to all postsynaptic ones. The synaptic memory has a word width of $J \times Q_{s}$, with $J$ the number of post-synaptic neurons and $Q_{s}$ the signed fixed-point precision of synaptic parameters. This memory organization ensures that the membrane potential of each NPU can be updated in parallel after one pre-synaptic spike. 
%all parameters required to update the membrane potential after one pre-synaptic spike can be retrieved with one clock cycle and redirected to the input of each NPU in parallel. 

Unlike Gyro, which implements a leaky-integrate-and-fire neuron model as an NPU, our architecture implements the dendrite-enhanced neuron model proposed in Section~\ref{sub:proposed_nn}. To achieve this goal, we propose a novel NPU, illustrated in \figrefsub{fig:fig3}{a}, and incorporate an additional memory module, named \textit{DENDRITE MEMORY}, for storing the dendritic delay of each NPU in a layer. The dendrite memory has a depth equivalent to the number of tasks, where each address has a width of $J \times Q_{d}$, with $Q_{d}$ being the unsigned fixed-point precision of the dendritic delay. This memory organization ensures that the dendritic delays for all neurons can be loaded concurrently. Specifically, when a layer receives the \textit{TASK\_ID} indicating the current task along with a \textit{LOAD\_TASK} signal, the main controller triggers the memory controller to access the dendrite memory. The memory output is then directed to the delay input port of each NPU and loaded into the corresponding down counter.

\section{Results}

\subsection{Software simulations}
\begin{figure}[ht!]
  \centering
  {\includegraphics[width=0.48\textwidth]{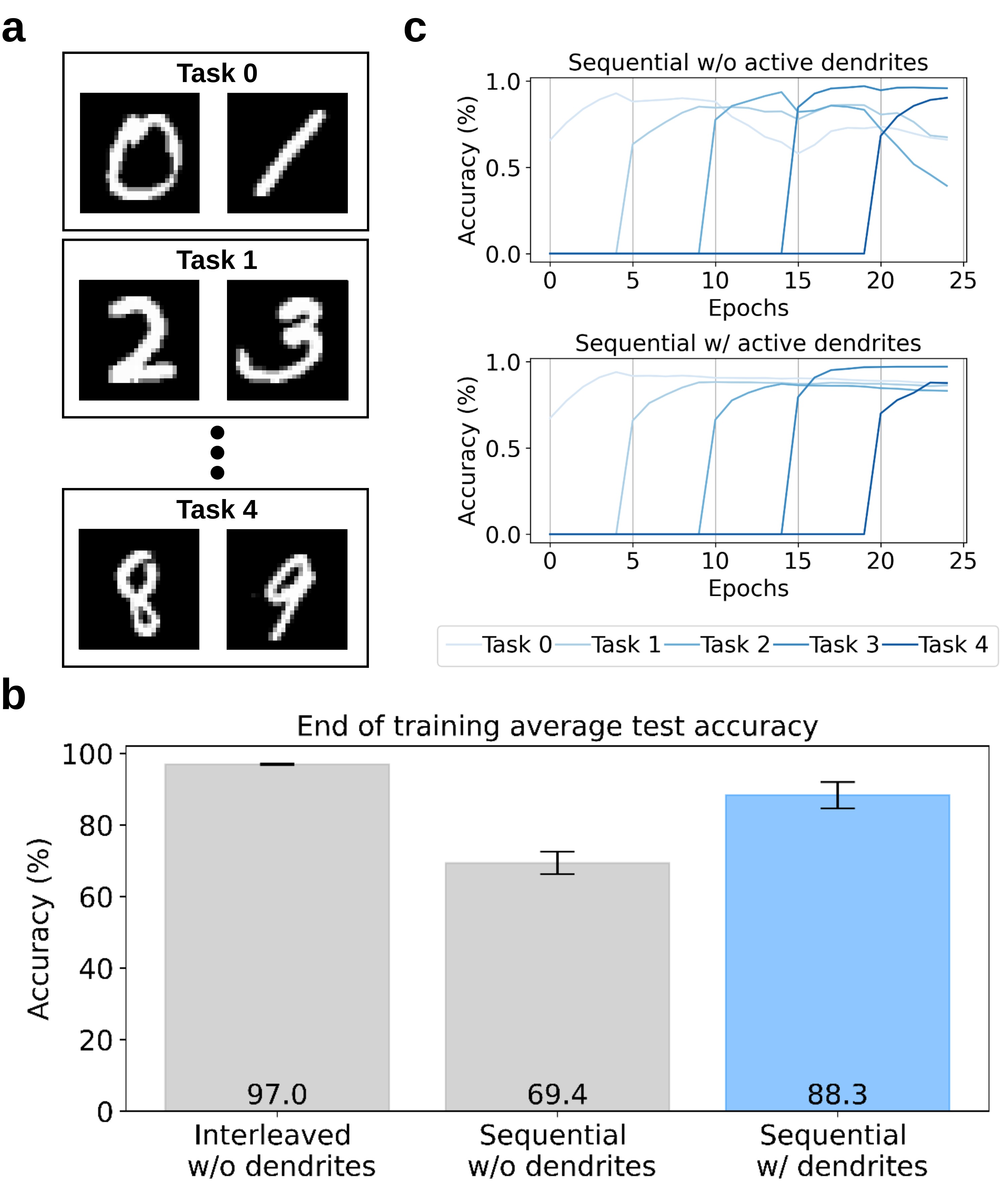}}
  \caption{\textbf{Split-MNIST setup and results.} \textbf{a} Example of three distinct tasks, each aiming to differentiate between two digits. \textbf{b} Test accuracy for each task over training time for the model without active dendrites (top) and with active dendrites (bottom). Note that a new task is introduced every 5 training epochs. \textbf{c} Average accuracy across tasks at the end of training~\mbox{for three experiments.}\vspace*{-2mm}} %The numbers within parentheses indicates the number of hidden neurons, thereby keeping the parameter size constant across experiments. }
  \label{fig:fig4}
\end{figure}

Our model's performance was evaluated in a single-head task-incremental scenario using the Split MNIST dataset, a popular benchmark for continual learning~\cite{2022_van_three}. It consists in sequentially training a neural network to solve 5 different tasks. Each task requires the network to discriminate between two consecutive digits of the MNIST dataset, e.g. 0 and 1, 2 and 3, etc. (\figrefsub{fig:fig4}{a}). Each digit is temporally encoded by transforming the pixel intensities $I_i$ into spike times as
$ t^{input}_{i} = T_{max}(I_{max}-I_i)/I_{max}$, where $I_{max}=256$ and $T_{max}=450$. 
%\figurerefsub{fig:fig4}{b} provides an example of a zero digit encoded in TTFS.
To evaluate the effectiveness of our solution, we conducted three experiments: 
\begin{enumerate}
    \item\textit{Interleaved without dendrites}: Establishing the upper performance bound using a TTFS-encoded network  with interleaved task presentation (i.e., no catastrophic forgetting).
    \item\textit{Sequential without dendrites}: Establishing the lower performance bound on the same TTFS-encoded network but with sequential task presentation (i.e., catastrophic forgetting).
    \item\textit{Sequential with dendrites}: Incorporating active dendrites in the TTFS-encoded network and presenting tasks in sequential order.
\end{enumerate}

The network architecture of experiments (1) and (2) is 784-403-403-2, while the network architecture in experiment (3) is 784-400-400-2, with all neurons in the hidden layers enhanced with active dendrites. The additional neurons in the hidden layers of the first two experiments ensure that all experiments have an equivalent number of learnable parameters. Similarly to previous works~\cite{2022_van_three}, we use the average test accuracy across all tasks at the end of training as a performance metric. All experiments use the Adam optimizer with a learning rate of 3e-4 for all learnable parameters and $S = 4$.

Each experiment was repeated for 5 different seeds and the results averaged. A summary of the performance for each experiment is provided in \figrefsub{fig:fig4}{b}. Experiment (3) shows a reduction of 8.7 accuracy points from the upper bound, while experiment (2) shows a more substantial reduction of 27.6 accuracy points. For a clearer visualization of the proposed solution's effectiveness in mitigating catastrophic forgetting, in \figrefsub{fig:fig4}{c} we plot the test accuracy for each task over the training duration. This figure illustrates the capability of the model enhanced with active dendrites to retain information on previous tasks as new tasks are being added. 

%To compensate for the additional capacity introduced by active dendrites, the hidden layers of experiments (1)-(2) are double the size of that of experiment (3), thereby keeping the model size constant. All experiments use a 2-hidden layer network. %Tasks in experiments (2)-(3) are trained sequentially for 5 epochs each. In experiment (1), where tasks are interleaved, the network is trained for 25 epochs. 

%Notably, despite experiment (2) having double the number of trainable parameters than experiment (3), the former achieves a lower test accuracy at the end of training than the latter. Precisely, experiment (3) shows a reduction of 8.7 accuracy points from the upper bound, while experiment (2) shows a more substantial reduction of 27.6 accuracy points.

%This suggests that our model can mitigate the problem of catastrophic forgetting in TTFS-encoded neural networks, without exclusively relying on an increased number of parameters. 

%Precisely, for an equal number of neurons between two hidden layers, the number of neurons required for a model without dendrites to match the parameter count of a model with active dendrites is $ \approx \sqrt{H_{w}^2+N H_{w}}$, where $H_w$ is the number of neurons in the model with dendrites. For instance, a model with 10 tasks and 2 hidden layers of 400 neurons each, the equivalent model without dendrites has only 405 neurons in each layer, indicating an efficient increase in the model parameters count. 

\subsection{FPGA implementation}
The proposed hardware architecture is implemented in the programmable logic of a Xilinx Zynq-7020 SoC FPGA. For monitoring, controlling and configuration purposes, the hardware architecture is connected to the processing system (PS) of the SoC using two AXI buses. The AXI buses are controlled from a Python environment running on the PS. Once an output spike is generated, the output address and spike time is written back to the memory of the PS. 

To limit the required on-chip memory, the synaptic weights and dendritic delays need to be quantized. Following full-precision training on a GPU cluster, the synaptic weights of the experiment (3) are quantized to 4-bit signed fixed-point, that is, $Q_s=4$ while the dendritic delays are quantized to 8-bit unsigned fixed-point, that is $Q_d=8$. We use 11-bit signed fixed-point to represent the membrane potentials of the neurons. We deploy the quantized network in both software simulation and in the FPGA. For the Split MNIST dataset, the FPGA implementation and the software simulation both achieve 80.0\% accuracy. The FPGA implementation matches the software simulation for all tasks and all samples, showcasing an average inference time of 37.3 ms for each image. Our design uses 93.2\% of LUTs, 35.3 \% of flip-flops and 29.3 \% of BRAMs.

\section{Conclusions}
In this paper, we have introduced a novel TTFS-encoded SNN model enhanced with active dendrites, which can mitigate the problem of catastrophic forgetting. We demonstrated competitive performance on the standard Split MNIST dataset, showcasing an end-of-training accuracy of 88.3\% across all tasks. Specifically, the network enhanced with active dendrites shows a reduction of only 8.7 accuracy points from the upper bound, while the same model without active dendrites shows a reduction of 27.6 accuracy points. Additionally, we proposed a novel digital hardware architecture that paves the way toward the deployment of continual-learning devices at the edge. Our proposed architecture has an average inference time of 37.3 ms and a test accuracy of 80.0\% when deployed on a Xilinx Zynq-7020 SoC FPGA. 

\section*{Acknowledgment}
This publication is funded in part by the project NL-ECO: Netherlands Initiative for Energy-Efficient Computing (with project number NWA. 1389.20.140) of the NWA research programme Research Along Routes by Consortia which is financed by the Dutch Research Council (NWO).  
\balance
\bibliographystyle{./bibliographies/IEEEtran}
\bibliography{./main.bib}
\end{document}